\documentclass[letterpaper, 10 pt, conference]{ieeeconf}  

\IEEEoverridecommandlockouts                              

\usepackage{graphics} 
\usepackage{epsfig} 
\usepackage{mathptmx} 
\usepackage{times} 
\usepackage{amsmath} 
\usepackage{amssymb}  
\usepackage{hyperref}
\usepackage{booktabs}
\usepackage{mathtools}
\usepackage{graphicx,tabularx}
\usepackage{svg}
\usepackage{multirow}
\usepackage{blindtext}  

\usepackage{subcaption}

\usepackage{xcolor}

\usepackage{soul}

\title{\LARGE \bf Flocking behavior for dynamic and complex swarm structures }

\author{Carmen DR.Pita-Romero, Pedro Arias-Perez, Miguel Fernandez-Cortizas, Rafael Perez-Segui, Pascual Campoy \\
Computer Vision and Aerial Robotics group (CVAR), Centre for Automation and Robotics (CAR) \\ Universidad Politécnica de Madrid (UPM-CSIC),  Madrid, Spain. 
\thanks{Corresponding author: c.derojas@alumnos.upm.es }
}

\begin{document}

\maketitle
\thispagestyle{empty}
\pagestyle{empty}

\color{black}
\begin{abstract}
Maintaining the formation of complex structures with multiple UAVs and achieving complex trajectories remains a major challenge. This work presents an algorithm for implementing flocking behavior of UAVs based on the concept of Virtual Centroid to easily develop a structure for the flock. The approach builds on the classical virtual-based behavior, providing a theoretical framework for incorporating enhancements to dynamically control both the number of agents and the formation of the structure. Simulation tests and real-world experiments were conducted, demonstrating its simplicity even with complex formations and complex trajectories.
\end{abstract}

\section*{\small{SUPLEMENTARY MATERIAL}}
\small
\noindent Video of the experiments: \url{https://vimeo.com/cvarupm/flocking-behavior} \\
Released code: \url{https://github.com/carmendrpr/project_flocking_behavior}
\normalsize
\color{black}
\section{INTRODUCTION}
Unmanned Aerial Vehicles (UAVs) are an emerging group of robots whose technology has advanced significantly in recent years. Their variety of sizes, designs, communication networks, scalability, and autonomous navigation capabilities offers exceptional flexibility in a wide range of domains \cite{surveyUAVs}.

Multi-Robot Systems (MRS) exploit the collective speed and maneuverability of multiple UAVs. Researchers \cite{MULTI_VS_SINGLE} have demonstrated both theoretically and empirically that MRS outperforms a single UAV in completing a task, both in terms of physical capabilities and execution time. However, this comes with an increase in the complexity of the system, especially regarding how to coordinate the different agents.

One of the many advantages of multi-UAV systems is their resilience to failure. If one agent fails, the others can take over its responsibilities to achieve the common goal. This capability increases the overall robustness of the system. Within the wide range of applications, it is important to emphasize their benefits in tasks that require the inspection of large areas of terrain in a limited time frame.

The organization of agents in an MRS is occasionally inspired by natural phenomena. Their formation control is based on the structures observed in biological colonies, such as flocks of birds, schools of fish, or the movement of geese. The research uses the term \textit{flocking} when referring to a type of collective behavior in which individual agents coordinate their movements to achieve a common task \cite{StateofArtFlocking}. In the context of a multi-UAV system, it is generally defined as bio-inspired collective behavior in which multiple unmanned aerial vehicles autonomously coordinate their movements to achieve a common goal.

The complexity of achieving complex trajectories with complex flocking structures is our main concern along with the scalability of the flock. Consequently, for this work we develop an algorithm to easily design an organized structure from a set of multi-UAV systems.


\subsection{RELATED WORK} 
Research on the design of flocking formations has identified two complementary approaches to their implementation \cite{olfati2006flocking, StateofArtFlocking}. From a theoretical perspective, the multi-UAV system must follow specific rules defined by Craig Reynolds to ensure collective motion \cite{reynolslaws}. From a physical perspective, the system must implement a well-defined control formation.

\subsubsection{Theoretical Perspective}
Craig Reynolds developed the boids algorithm in 1987 \cite{reynolslaws}, which simulates the flocking behavior of birds.
This algorithm establishes three basic rules that provide each agent with a guideline for evolving behaviors that allow them to integrate into a flock and generate coordinated collective motion.
These rules are defined as:

\begin{enumerate}
    \item \textbf{Cohesive Behavior:} assures adherence to the flock, each agent will gravitate toward the center of the herd, maintaining group unity and ensuring collective movement.
    \item \textbf{Separation Behavior:} ensures agents maintain a safe distance from each other to avoid collisions and prevent overcrowding within the swarm.
    \item \textbf{Alignment Behavior:} involves adjusting the agent's speed to match the speed and direction of its neighbors, ensuring synchronized movement and coordination within the herd.
\end{enumerate}

\subsubsection{Physical Perspective}
According to \cite{StateofArtFlocking}, a survey conducted on the topic of flocking behavior for multi-robot systems, the design separates it into task distribution and flocking control.

Based on task distribution, multi-robot systems are classified into two groups: heterogeneous systems, where each agent is assigned a different task, and homogeneous systems, where all agents perform the same task using the same strategy \cite{StateofArtFlocking}.


The flocking control divides this taxonomy into three components: \textit{schemes}, \textit{structures}, and \textit{strategies} \cite{StateofArtFlocking, algorithmscontrol, miguel}. 

\begin{itemize}
    \item \textbf{Schemes:} refer to the type of framework used. It includes defining how knowledge is shared among all agents, as well as the communication protocols between them. The scheme is classified as distributed, decentralized, centralized, or hybrid systems \cite{StateofArtFlocking}. In a distributed scheme, knowledge is shared, and each agent is controlled independently \cite{distributed, distributed_2}. Decentralized systems, on the other hand, lack direct communication between agents \cite{descentralized, mejias2024virtual}. Centralized schemes involve a central unit to which all members are connected and controlled  \cite{centralized}. The hybrid system combines the distributed scheme with the centralized approach, providing both knowledge availability and reduced complexity in terms of time and space \cite{pinning}. 
    
    \item \textbf{Structures:} are responsible for organizing the distribution of agents within the available space. It also specifies how their movement is coordinated and outlines the mechanisms for achieving scalable flocking behavior.
    
    Classically, these structures are categorized into three main approaches: Leader-Follower, Behavior-Based and Virtual-Based. The Leader-Follower approach is particularly well suited to maintaining a hierarchical structure while achieving simplicity and scalability \cite{leader_follower, leader_follower2}. In this model, the leader agent determines the behavior and trajectory, while the follower agents follow the leader's path. 
    
    However, this approach has significant drawbacks. Due to its high dependence on the leader, the structure becomes vulnerable to failure if the leader malfunctions or fails, significantly reducing its robustness. In addition, this approach is not well suited for achieving geometric formations among agents. Behavior-based systems rely on predefined behaviors, such as cohesion, obstacle avoidance, and collision avoidance, to guide agents to achieve collective goals through local interactions \cite{behavior-based}. These systems emphasize decentralized decision making, allowing agents to adapt to dynamic environments while maintaining group integrity and task efficiency. Virtual-based systems rely on virtual links or reference points to coordinate agents, using a virtual leader to guide movements and effectively maintain geometric formations \cite{virtual, virtual2, VirtualStructure}. In recent years, new techniques derived from the classical ones have been considered. Pinning-based structures adhere to the leader-follower model, but by pinning multiple agents, they increase stability \cite{pinning}.

     \item \textbf{Strategies:} involve guiding the movement of agents in a swarm to maintain desired formations while avoiding obstacles. Traditional methods, like Reynolds' rules and Artificial Potential Fields (APF) \cite{PotencialField}, focus on behaviors such as cohesion and collision avoidance. Modern approaches, including swarm intelligence \cite{swarminte} and AI-based techniques, use bio-inspired algorithms \cite{bio} and machine learning \cite{matching} to improve flexibility and efficiency. Hybrid control paradigms combine multiple strategies to improve performance and robustness in dynamic environments.
\end{itemize}

\begin{figure}[h]
    \centering
    \includegraphics[width=1\linewidth]{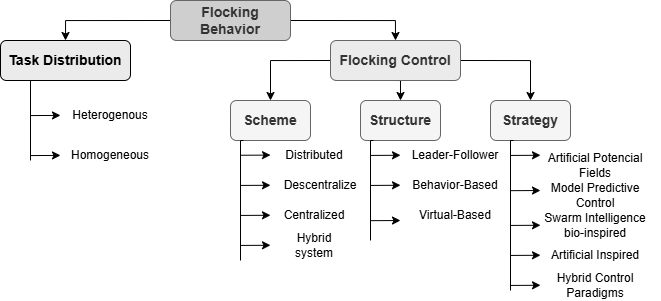}
    \caption{Taxonomy of flocking behavior in multi-agent systems.}
    \label{fig:flocking}
\end{figure}

Fig. \ref{fig:flocking} provides a structured guideline for conducting flocking behavior in multi-agent robotic systems.

\subsection{CONTRIBUTION}
In this paper, we present \textit{FlockingBehavior}, an algorithm for constructing complex structures for $N$ agents that can easily perform complex trajectories. This algorithm is inspired by the virtual-based approach. The experiments conducted use a centralized scheme and an offline control strategy. However, it can be used with any scheme and strategy desired. The main contributions of this work are as follows:

\begin{enumerate}
    \item The formalization of an algorithm that allows the user develop arbitrarily complex structures with no limitations on the number of agents. The algorithm is characterized by the simplicity to modify the organization of the agents inside the swarm and to restructure the formation in case of necessity to add or remove new agents. 
    \item Extensive simulation experiments and a proof-of-concept test with real drones have been performed, validating the proposed algorithm. The experiments carried out utilizes geometric rigid formations.
    \item The code is open source, aiming to contribute to the scientific community, allowing researchers to evaluate and build upon it. Its design follows the modular philosophy of ROS 2.
\end{enumerate}

\color{black}

\section{FLOCKING BEHAVIOR FOR UAVs}
The overall problem addressed in this work is how to design a flocking behavior, allowing a set of agents to develop an organized structure that guarantees satisfying the Reynolds' rules. There is no limit to the complexity of the structures, and flights can be performed with dynamic geometric formations. The goal is to simplify the task of performing formation flights by eliminating the need to define an individual trajectory for each agent, one of the main constraints to ensure alignment during the collective motion of the flock.

\subsection{\textbf{Problem Formulation}}
We can formulate the problem in the following terms. Given a set of $N$ agents, the objective is to find the function $\mathbf{F}$ which provides a reference pose $P_i^W(t)\in SE(3)$ for each agent $i$ ---in the inertial frame $W$--- so that the virtual centroid of formation ($VC$) follows an input trajectory $\sigma_{VC}^W(t)$---in the frame $W$--- while keeping a predefined geometric figure $G_i^{VC}(t,N)$ with respect to the virtual coordinate system $VC$.
\begin{align}
P_i^W(t) = \mathbf{F}\left(\sigma_{VC}^W(t),G_i^{VC}(t,i)\right)\quad ;\forall i\in\{0,N\}\label{eq:p_form}
\end{align}

In this work, we will make the following assumptions, in order to develop the structure control algorithm, and to validate it:
\begin{enumerate}
    \item \emph{Feasible Trajectories}: we consider that the trajectories generated $P_i^W(t)$ are feasible for the dynamic constraints of the agents and has a maximum speed of $V_{max}$.
    \item \emph{Precise control}: we consider that each agent $i$ is capable of following the generated references $P_i^W(t)$ precisely.
    \item \emph{No obstacle in the environment}: we consider that the swarm formation moves within the free space.
\end{enumerate}




The generated swarm structure shall be consistent with Reynolds’ laws, which are fundamental principles governing the behavior of flocking systems. Taking into account the former assumptions, we can represent these laws as follows:
\vspace{-0.7cm}
\begin{align}
\intertext{Cohesion:}
&||P_i^W(t),P_{VC}^W(t)|| < d_{max}\qquad &;&\;\forall i,j\in\{1,N\}\label{eq:cohe}\\
\intertext{Separation:}
&||P_i^W(t),P_j^W(t)|| > d_{min}\qquad &;&\;\forall i,j\in\{1,N\}|i\neq j\label{eq:sepa}\\
\intertext{Alignment:}
&||\dot{P}_i^{VC}(t)|| < \delta &;&\;\forall i\in\{1,N\}\label{eq:align}
\end{align}

Where $d_{min}>0$ the minimum distance between agents, $d_{max} > 0$ the maximum distance from an agent to the center of the formation, and $\delta>0$ the maximum difference in the relative speed of each agent. $||v||$ represents the norm $2$ of the vector $v$, and $||v,w||$ the norm $2$ of the vector formed from $v$ to $w$.



Below, the algorithm proposed is presented and the particularities of how to deploy this flocking behavior in a swarm.

\subsection{\textbf{Structure generation}}
From the problem formulation, we need to further specify what each term means and how it can be formulated for accomplishing the desired Reynolds rules. First of all, we define:
\begin{equation}
    \sigma_{VC}^W(t) \in SE(3)
    = \begin{bmatrix}
    R_{VC}^W(t) & T_{VC}^W(t)\\
    0 &1
    \end{bmatrix} 
\end{equation}

as the transformation between the inertial frame $W$ with the origin of the coordinate system $VC$ fixed to the swarm formation centroid. $R(t)\in SO(3)$ represents the rotation of the swarm formation and $T(t)\in\mathbb{R}^{3\times1}$ refers to the translation of the swarm formation. 

Then $G_i^{VC}(t,i) \in SE(3)$ represents the desired formation, in terms of the transformation between the centroid of the formation and a specific agent. The use of these representations of the geometric figure in terms of $SE(3)$ transformations allows us to compute the reference poses trivially by:
\begin{align}
P_i^W(t) = \sigma_{VC}^W(t)G_i^{VC}(t,i)
\end{align}

\begin{figure}[!htb]
    \centering
    \includegraphics[width=0.8\linewidth]{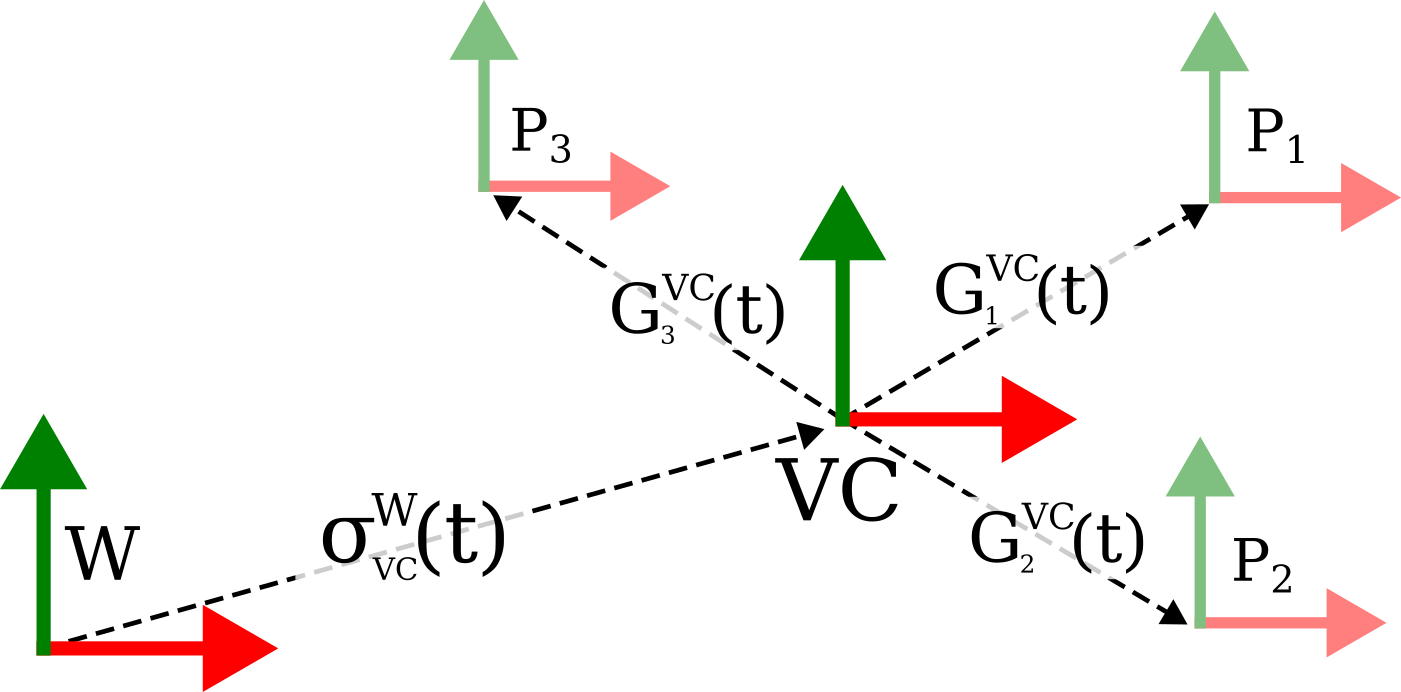}
    \caption{Coordinated frames of the inertial frame W, the virtual centroid VC and each agent of the swarm and their relationships.}
    \label{fig:frames_flocking}
\end{figure}

In Fig \ref{fig:frames_flocking}, we can see these reference frames and poses in a three-agent example.

With this approach, given a trajectory to the centroid of the swarm, we can generate the trajectory that each drone shall follow, based on the geometric figure. Thus, the most important part is setting up how this geometry is defined and how to preserve the compliance of the Reynold's laws.

In the following, we further explain how the structure is set and how the coordinated movement of the swarm is obtained.

\subsection{\textbf{Setup the structure}}

The objective of the setup phase is to define the formation of the swarm and the initial pose desired for its centroid prior to starting the collective motion. The setup phase corresponds to the transition from each agent's initial pose at $t_0$ to its positioning in the formation at $t_s$.

To achieve this, we need the pose of the virtual centroid $P_{VC}^W(t_s)$ to define the origin of the swarm at the setup time $t_s$, and the initial formation geometry for the $N$ agents $G_i^{VC}(t_s, i)$.
    
\begin{figure}[h]
    \centering
    \includegraphics[width=1\linewidth]{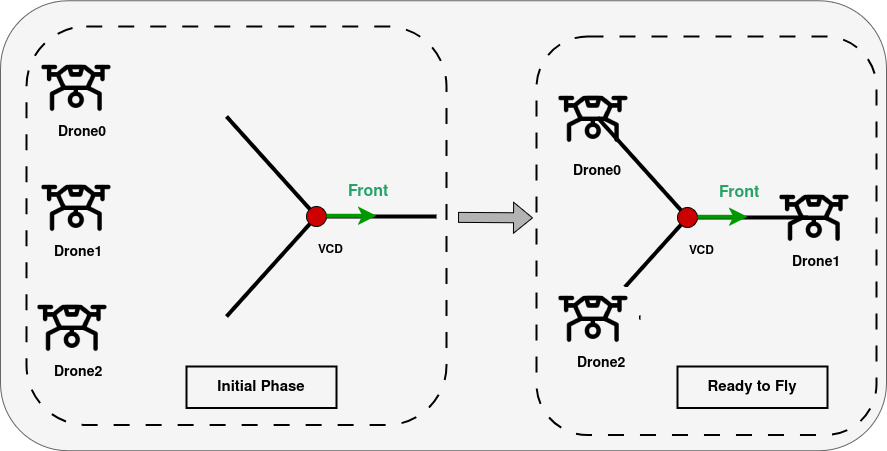}
    \caption{Overview of the setup phase. On the left, it is pictured time $t_0$ prior to the formation to be shaped. On the right, UAVs occupy their position ready to start the collective movement at $t_s$. Red dot represent the virtual centroid, while green arrow points to its front orientation.}
    \label{fig:strucuture}
\end{figure}

During the set up, the geometry is modeled as a rigid body. The relative position of the $VC$ with respect to the global frame is defined in by the trajectory $\sigma_{VC}^T(t)$. 
As an example, planar static regular formations with $N$ drones with a $d_{max}$ could be generated by using the following function:
\begin{align}
    &G^{VC}_i(d_{max},N)= \begin{bmatrix}
        R & T_i \\ 0 & 1
    \end{bmatrix} &;&\;
    R = I^{3\times3};\\
    &T_i = \begin{bmatrix}
        d_{max} \cdot cos\left(\theta\right)\\
        d_{max} \cdot sin\left(\theta\right)\\
        0
    \end{bmatrix}  &;&\; \theta = \frac{2\pi i}{N} \label{eq_t}\\
    \intertext{constrained to :}
    &d_{max}\cdot2 \cdot sin\left( \frac{N}{\pi} \right) \geq d_{min} && \label{eq_const}
\end{align}
The previous geometry formation follows the heading of the $VC$ so that the rotation matrix $R$ is the Identity. Fig. \ref{fig:strucuture} shows a three-drone formation using the previous geometry distribution example.

When opting for static formation, we can model the swarm as a rigid body, which trivially guarantees the Reynold's laws. In the former example, as the formation is static, we can consider that each agent does not move with respect to the $VC$ so $\dot{P}_i^{VC}(t) = \vec{0} \rightarrow ||\dot{P}_i^{VC}(t)|| = 0$ satisfying the alignment rule in eq. \ref{eq:align}.

Regarding the cohesion and separation laws, the rigid body assumption guarantees that the distance between two points in the formation, which can be two different agents or one agent with respect to the centroid, will not vary over time. When using the former example, the constraint $d_{max}$ is considered as one of the inputs, so the formation is designed with the separation between the origin and each agent. Then, following eq. \ref{eq_t}, cohesion rule can be understood as $||T_i|| = d_{max}$. Finally, in order to ensure the separation rule, we need to add the constraint in eq. \ref{eq_const} in the former algorithm.

Although in this work we will use static formations during collective motions, we could change the formation over time or between motions, gaining the ability to modify the swarm size and reorganize the swarm formation.



With this, we can generate the starting reference poses $P_i^W(t_s)$ for each agent, then the UAVs navigate to their position with the desired orientation, finalizing the setup phase. Next section describes how the collective motion is achieved while preserving the structure of the formation.

\subsection{\textbf{Dynamic structure during collective motion}}

Once the swarm formation has been established, the objective is to ensure that the structure is maintained at all times, regardless of whether the structure is modified during flight or continuously adjusted.

The trajectory following of the virtual centroid and the frame following of each agent in the swarm is what generates this emergent collective motion. The pose of the virtual centroid $\mathbf{p}_{VC}^W$ evolves over time, which can be generally assumed as the evaluation of a trajectory. Notice that the UAVs are not given a regenerated trajectory; on the contrary, their $P_i^W(t)$ is computed online based on $\sigma_{VC}^W(t)$, triggering the movement of the UAVs. From this point, the control of the algorithm is open-loop. 


One of the advantages of this approach is that all agents within the swarm are treated with equal importance. The agents depend only on the trajectory of the $VC$ which centralizes the formation. 

As a contribution to the classical virtual structure-based approach, the algorithm allows changing the structure of the swarm $G_i^{VC}(t,i)$. So the swarm can operate with both rigid structures and dynamic structures that can vary over time or be updated according to a specific model.

The algorithm also allows the number of agents in the formation to change during execution. To handle such situations, the algorithm implements methods to add or detach agents to the swarm, if they satisfy the constraint of $d_{min}$ between them. 


These two contributions provide significant flexibility within the structure definition, allowing the formation to be modified and the number of agents to be dynamically adjusted.

It is worth mentioning that the swarm formation has an orientation, it means that it ``faces'' a direction.
This allows the swarm to determine which part of the formation should always face the direction of motion, allowing it to rotate the entire structure and smoothly execute curved trajectories. 



\section{EXPERIMENTAL VALIDATION}
To evaluate the performance of our algorithm, we conducted a series of experiments. The metrics are carried out following the proposed formulas for cohesion eq. \ref{eq:cohe}, separation eq. \ref{eq:sepa} and alignment eq. \ref{eq:align} according to the explication on the \textit{FlockingBehavior} to validate the adherence to the Reynolds laws.

We evaluate the algorithm's ability to execute trajectories with path facing. In addition, we evaluate its scalability in handling varying numbers of agents, allowing for in-flight modifications, and adjusting the formation accordingly.
The experimental validation was performed in both real and simulated environments with different swarm sizes and formations.

Results shown in this section correspond to steady-state flight only to avoid discrepancies. Pruning data collected during the experiments prevents errors from the takeoff, landing, and positioning of the drones in their $P_i^W(t_s)$ during the setup phase.

\subsection{Experimental setup}
The UAVs used in our experiments are Bitcraze Crazyflie 2.1 quadcopters, while a dynamic model simulator is used for simulation. Aerostack2 \cite{Aerostack2} is chosen as the aerial robotics framework. We rely on the Aerostack2 control and communication with the Crazyflies. To generate and evaluate the trajectory to be followed by the swarm, we use a dynamic polynomial 3D trajectory offered in the Aerostack2 framework.

\begin{table*}[h]
    \centering
    \begin{tabular}{cc|c|c|ccc|c}
    \hline
    \hline
     & \multirow{2}{6em}{\textbf{$v_{swarm}$ [m/s]}} & \multirow{2}{6em}{\textbf{Cohesion [m]}} & \multirow{2}{6em}{\textbf{Reference Error [m]}} &\multicolumn{3}{c|}{\textbf{Separation [m]}} & \multirow{2}{7em}{\textbf{Aligment [m/s]}} \\
     &  &  && \textit{drone0} & \textit{drone1} & \textit{drone2} & \\
    \hline
	\textit{drone0} & & 0.849 ± 0.367 & 0.238 ± 0.189&- & 2.262 ± 0.066& 2.309 ± 0.069 & 0.057 ± 0.066\\     
	\textit{drone1} & 0.5 &1.662 ± 0.182 & 0.235 ± 0.187 &2.262 ± 0.066 & - & 2.022 ± 0.015 & 0.062 ± 0.056 \\    
	\textit{drone2} & & 1.663 ± 0.181 & 0.239 ± 0.188&2.309 ± 0.069 & 2.022 ± 0.015 & - & 0.064 ± 0.063 \\
    \hline
	\textit{drone0} & & 0.846 ± 0.363 &0.239 ± 0.189 &- & 2.206 ± 0.048 & 2.291 ± 0.063 &0.059 ± 0.066 \\
	\textit{drone1} & 1 & 1.662 ± 0.181 & 0.238 ± 0.188 &2.206 ± 0.048 & - & 2.026 ± 0.018 & 0.062 ± 0.066 \\
	\textit{drone2} & & 1.657 ± 0.178 & 0.235 ± 0.184& 2.291 ± 0.063 & 2.026 ± 0.018 & - & 0.060 ± 0.064 \\
    \hline
	\textit{drone0} & & 0.844 ± 0.357 &0.238 ± 0.187 &- &2.223 ± 0.037 & 2.246 ± 0.036& 0.067 ± 0.068\\
	\textit{drone1} & 2 & 1.658 ± 0.179 & 0.236 ± 0.186&2.223 ± 0.037 & - & 2.017 ± 0.010 & 0.065 ± 0.063 \\
	\textit{drone2} & & 1.658 ± 0.178 & 0.239 ± 0.187&2.246 ± 0.036 & 2.017 ± 0.010 & - & 0.068 ± 0.067 \\
    \hline
    \hline
    \end{tabular}
    \caption{We report the average and standard deviation for the metrics to compare cohesion, reference error, separation, and alignment in the linear trajectory.}
    \label{tab:linear}
\end{table*}

\subsection{Simulated Scenarios}
The chosen swarm structure consists of three drones in an equilateral triangle formation. The corresponding parameters for the structure are $2 m$ for the side and $1.414 m$ between the $VC$ and the drones. The drones are placed in the environment, and the demanded structure is set up before the flight. We evaluated the algorithm through several experiments and repeated every experiment ten times to reduce the stochastic effects of the simulation. 

\subsubsection{Linear Trajectory}
The goal is to perform a 15 meter linear trajectory while maintaining the equilateral triangle formation within the swarm in an obstacle-free scenario. We performed ten simulation runs, varying the maximum speed of the swarm at $0.5 m/s$, $1 m/s$, and $2 m/s$. 

\begin{figure}[h]
    \centering
    \includegraphics[width=0.9\linewidth]{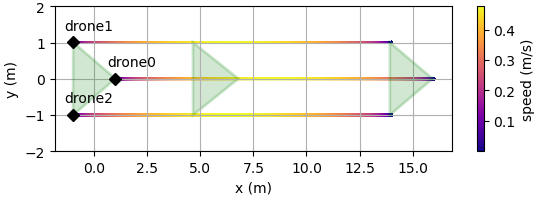}
    \caption{Zenithal view with the linear trajectory according to the individual velocity of each drone. Color map on the right represents velocities between 0.0 and 0.5 $m/s$. The colored triangle represents the formation of the swarm at the beginning, middle, and end of the trajectory.}
    \label{fig:trayectoria_lineal}
\end{figure}

Fig. \ref{fig:trayectoria_lineal} shows the zenithal view of the trajectory according to the individual speed of each drone to show the alignment of the swarm. The colored triangle represents the formation of the swarm at the beginning, middle, and end of the trajectory, showing that the formation is preserved throughout the flight.

Table \ref{tab:linear} presents a summary of the mean and the standard deviation of the experiments divided by the set speed of their swarm.

The results in the cohesion metrics show that two drones are farther from the centroid than the one ahead. This is a consequence of the geometric formation chosen; one drone is in front of the $VC$, while two are behind it. As they move forward and lag slightly behind their references, the leading drone aligns more closely with the $VC$ than the others. The reference error $RE$ quantifies this discrepancy.
\begin{equation}
    RE = \int_{t_s}^{t_{end}} ||P_i^W(t),P_i^{VC}(t)||\qquad ;\forall i\in\{0,N\}
    \label{eq:reference error}
\end{equation}

Taking a look at the result for $v_{swarm}=0.5 m/s$, \textit{drone0} is ahead of $VC$. Its cohesion metric shows a value $0.849 \pm 0.367 m$ and a reference error metric of $0.238 \pm 0.189 m$. A corrected cohesion metric can be obtained with the addition of both values, giving a result of $1.087 \pm 0.412 m$, which is similar to the expected value of $1.414 m$. 

The separation values observed between \textit{drone1} and \textit{drone2} slightly differ from the goal metric of $2 m$. Values obtained \textit{drone0} with their siblings are lightly bigger due to the direction of motion, as explained before during the cohesion discussion.

Following the definition of the alignment metric in eq. \ref{eq:align}, to maintain the trajectory throughout the flight, the velocities of the drones in the $VC$ frame must be close to zero $||\dot{P}_i^{VC}(t)|| = 0$. The values observed in the result table satisfy this requirement since they are around half a decimeter per second.

\begin{figure}[htb!]
    \centering
    \includegraphics[width=0.9\linewidth]{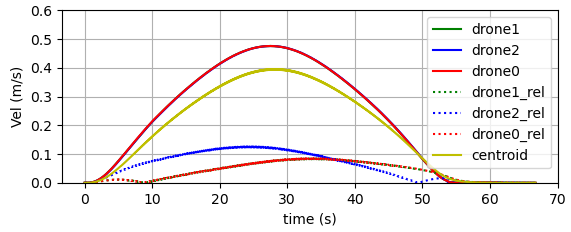}
    \caption{Speed for the linear trajectory experiment at $0.5m/s$. Solid curves represent each drone speed $||\dot{P}_i^{W}(t)||$, while dotted curves stand for their relative speeds to the centroid $||\dot{P}_i^{VC}(t)||$. Yellow curve shows the speed of the centroid $||\dot{P}_{VC}^{W}(t)||$.}
    \label{fig:Velocidad_lineal}
\end{figure}

Fig. \ref{fig:Velocidad_lineal} shows each drone speed $||\dot{P}_i^{W}(t)||$ during the experiment, together with their relative speed to the centroid $||\dot{P}_i^{VC}(t)||$ and the centroid speed $||\dot{P}_{VC}^{W}(t)||$. Both centroid and drone speeds exhibit a maximum value close to the reference value of $0.5m/s$. Moreover, the relative speed of the drones is below $0.14 m/s$, as reflected during the alignment metric analysis. Also, drone speeds are identical between them, since the three trajectories performed are also really similar. $Drone0$ relative speed differs from the other drones due to its position at the front of the formation ahead $VC$.

\subsubsection{Curvilinear trajectory}

\begin{table*}[!h]
    \centering
    \begin{tabular}{cc|c|c|ccc|c}
    \hline
    \hline
     & \multirow{2}{6em}{\textbf{$v_{swarm}$ [m/s]}} & \multirow{2}{6em}{\textbf{Cohesion [m]}} & \multirow{2}{6em}{\textbf{Reference Error [m]}} &\multicolumn{3}{c|}{\textbf{Separation [m]}} & \multirow{2}{7em}{\textbf{Aligment [m/s]}} \\
     &  &  && \textit{drone0} & \textit{drone1} & \textit{drone2} & \\
    \hline
	\textit{drone0} & &  0.862 ± 0.329  &  0.235 ± 0.135&- & 2.219 ± 0.031&2.274 ± 0.043 & 0.071 ± 0.063\\     
	\textit{drone1} & 0.5 &1.643 ± 0.160 & 0.235 ± 0.187 &2.219 ± 0.031 & - & 1.999 ± 0.028 & 0.059 ± 0.064\\    
	\textit{drone2} & & 1.649 ± 0.158& 0.239 ± 0.188&2.274 ± 0.043 &1.999 ± 0.028 & - & 0.067 ± 0.064 \\
    \hline
	\textit{drone0} & & 0.862 ± 0.337 &0.239 ± 0.138  &- & 2.168 ± 0.057 & 2.225 ± 0.038 & 0.061 ± 0.063  \\
	\textit{drone1} & 1 & 1.648 ± 0.162 & 0.201 ± 0.123&2.168 ± 0.057 & - & 2.004 ± 0.046 &0.063 ± 0.063\\
	\textit{drone2} & & 1.650 ± 0.160  &0.284 ± 0.170& 2.225 ± 0.038& 2.004 ± 0.046 & - & 0.070 ± 0.064\\
    \hline
	\textit{drone0} & &0.866 ± 0.334&0.234 ± 0.135 &- &2.205 ± 0.065 & 2.209 ± 0.029& 0.067 ± 0.063\\
	\textit{drone1} & 2 & 1.647 ± 0.162 & 0.200 ± 0.123 &2.205 ± 0.065& - &2.022 ± 0.051 & 0.064 ± 0.064 \\
	\textit{drone2} & & 1.652 ± 0.160 & 0.287 ± 0.171&2.209 ± 0.029 & 2.022 ± 0.051 & - & 0.072 ± 0.064\\
    \hline
    \hline
    \end{tabular}
    \caption{We report the average and standard deviation for the metrics to compare cohesion, reference error, separation, and alignment in the curvilinear trajectory }
    \label{tab:curvilinear}
\end{table*}

This experiment consists of executing a $20m$ length curvilinear trajectory maintaining an equilateral triangle formation in an obstacle-free scenario. We aim to prove the path-facing ability of the swarm while preserving Reynolds' laws.  We performed ten simulation runs, varying the maximum speed of the swarm at $0.5 m/s$, $1 m/s$, and $2 m/s$.

\begin{figure}[htb!]
    \centering
    \includegraphics[width=0.9\linewidth]{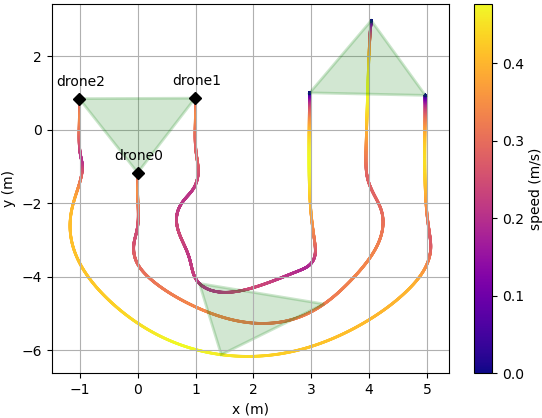}
        \caption{Zenithal view with the curvilinear trajectory according to the individual velocity of each drone. Color map on the right represents velocities between 0.0 and 0.5 $m/s$. The colored triangle represents the formation of the swarm at the beginning, middle, and end of the trajectory.}
    \label{fig:trayectoria_curva}
\end{figure}

Fig. \ref{fig:trayectoria_curva} shows how the swarm rotates while always keeping the formation facing the trajectory. The color map also shows how the drones adjust their speed based on their distance to $P_i^{VC}(t)$, but always below the set maximum speed allowed for each agent of $0.5 m/s$. 

The results obtained are summarized in Table \ref{tab:curvilinear}. Values collected for cohesion, separation, and alignment metrics verify the compliance of the Reynolds' laws. Indeed, results gathered are similar to the ones achieved in the linear trajectory, proving the robustness of our method against different trajectories.

\begin{figure}[htb!]
    \centering
    \includegraphics[width=0.9\linewidth]{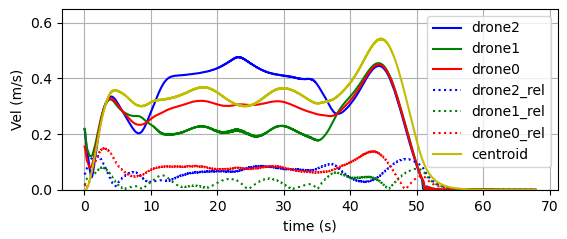}
    \caption{Speed for the curvilinear trajectory experiment at $0.5m/s$. Solid curves represent each drone speed $||\dot{P}_i^{W}(t)||$, while dotted curves stand for their relative speeds to the centroid $||\dot{P}_i^{VC}(t)||$. Yellow curve shows the speed of the centroid $||\dot{P}_{VC}^{W}(t)||$.}
    \label{fig:Velocidad_curva}
\end{figure}

Fig. \ref{fig:Velocidad_curva} shows each drone speed $||\dot{P}_i^{W}(t)||$ during the experiment, together with their relative speed to the centroid $||\dot{P}_i^{VC}(t)||$ and the centroid speed $||\dot{P}_{VC}^{W}(t)||$. \textit{drone2} is positioned outside the curve, so to maintain the formation, its speed is higher than the other drones. Similarly, \textit{drone1} has the lowest speed in the swarm, while \textit{drone0} has a similar speed to the centroid one due to its position in front. The relative speeds, as expected, are below $0.15 m/s$, ensuring that the alignment law is fulfilled.

\subsubsection{Dynamic Formation}
In order to prove the dynamic reconfiguration of the structure, we performed the following experiment. Having an initial structure of a $2m$ square swarm with four drones while flying a linear trajectory, one drone malfunctions, and the swarm reconfigures the structure into an equilateral triangle with side $2m$. Drone failure is simulated, and reconfiguration is done manually to demonstrate this capability.

\begin{figure}[htb!]
    \centering
    \includegraphics[width=0.9\linewidth]{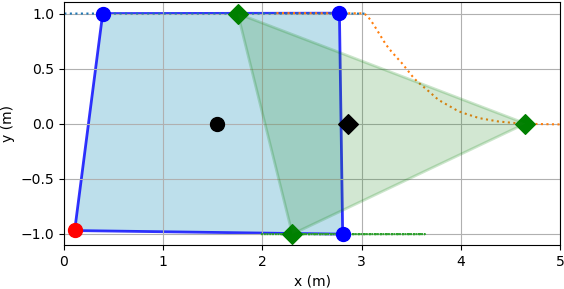}
    \caption{Swarm reconfiguration from a square (in blue) to an equilateral triangle (in green) formation. Dots are the drones at a time where the failure is detected and diamonds are the drones after the reconfiguration. Blue dots are the drones that stay in the swarm, while the red dot represent the drone that collapses. Black dot is the $VC$ in the initial stage and black diamond is the $VC$ after the reconfiguration. Green diamonds are the three drones in the final configuration.}
    \label{fig:dynamic_formation}
\end{figure}

Fig. \ref{fig:dynamic_formation} shows the moment when the swarm reconfigures and detaches $P_3^{VC}(t)$ from the structure. Time consumed for the reconfiguration is around $1.5 s$, showing the little time employed to achieve the new structure for the swarm without stopping the trajectory followed.

\subsubsection{Scalability of the algorithm }
To validate the scalability of the algorithm, a 3D trajectory was performed with a swarm of twelve drones. The video in the supplementary material shows the twelve drones performing the trajectory, maintaining the formation for this complex trajectory.
This simulation proves how easily the number of agents can increase and create a complex structure. Moreover, it proves that it doesn't matter how complex the trajectory is; our algorithm simplifies the task without the need to generate twelve specific trajectories.

\begin{table*}[!ht]
    \centering
    \begin{tabular}{cc|c|c|ccc|c}
    \hline
    \hline
     & \multirow{2}{6em}{\textbf{$v_{swarm}$ [m/s]}} & \multirow{2}{6em}{\textbf{Cohesion [m]}} & \multirow{2}{6em}{\textbf{Reference Error [m]}} &\multicolumn{3}{c|}{\textbf{Separation [m]}} & \multirow{2}{7em}{\textbf{Aligment [m/s]}} \\
     &  &  && \textit{drone0} & \textit{drone1} & \textit{drone2} & \\
    \hline
	\textit{drone0} & & 0.877 ± 0.440 &0.180 ± 0.152 &- & 1.153 ± 0.106 & - &0.105 ± 0.130 \\
	\textit{drone1} & Linear trajectory &1.001 ± 0.492 & 0.314 ± 0.138&1.153 ± 0.106 & - & -&0.112 ± 0.134\\
    
    \hline
\textit{drone0} & & 0.331 ± 0.461 &  0.124 ± 0.045&- & 1.064 ± 0.019&2.069 ± 0.057 & 0.037 ± 0.090 \\     
	\textit{drone1} & Z-axis rotation & 0.331 ± 0.461 & 0.124 ± 0.045 &1.064 ± 0.019 & - & 1.018 ± 0.067& 0.038 ± 0.089 \\    
	\textit{drone2} & & 1.213 ± 0.263&0.237 ± 0.088 &2.069 ± 0.057 &1.018 ± 0.067 & - & 0.038 ± 0.091 \\
    \hline
    \hline
    \end{tabular}
    \caption{We report the average and standard deviation for the metrics to compare cohesion, reference error, separation, and alignment for the real experiments.}
    \label{tab:real}
\end{table*}

\subsection{Real world experiments}
To validate our approach in real-world scenarios, we perform experiments varying the number of agents in the swarm and the type of motion. First, we perform a linear trajectory with a swarm of two drones forming a line separated by $1 m$, with the $VC$ positioned between the two drones. Once the drones are positioned, the swarm rotates to face the path and starts a $3 m$ linear trajectory. 

\begin{figure}[htb!]
    \centering
    \includegraphics[width=0.9\linewidth]{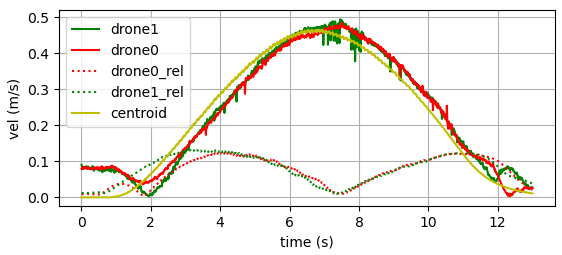}
    \caption{The solid lines are the absolute velocities for the swarm and the three drones. The dotted lines are the relative speeds of the drones in the real linear trajectory.}
    \label{fig:twist2}
\end{figure}
\begin{figure}[htb!]
    \vspace{-0.5cm}
    \centering
    \includegraphics[width=0.9\linewidth]{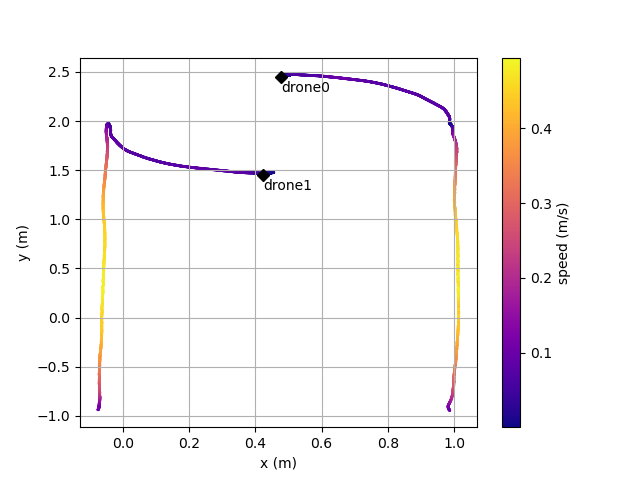}
    \caption{Zenithal view for the real experiment with the linear trajectory according to the individual velocity of each drone. Color map on the right represents velocities reached.}
    \label{fig:colored2}
\end{figure}

Fig. \ref{fig:colored2} shows the trajectory achieved by the swarm with its corresponding color map for the speed. Table \ref{tab:real} summarizes the results obtained.

Fig. \ref{fig:twist2} shows each drone speed $||\dot{P}_i^{W}(t)||$ during the experiment, together with their relative speed to the centroid $||\dot{P}_i^{VC}(t)||$ and the centroid speed $||\dot{P}_{VC}^{W}(t)||$. As expected, results obtained are very similar to the simulation one \ref{fig:Velocidad_lineal}. Furthermore, the relative speed of the drones is below $0.12 m/s$, as reflected during the alignment metric analysis.

The second experiment shows a swarm of 3 drones in a line formation performing a rotation along the z-axis. Each drone is separated from the others by $1 m$. The $VC$ is positioned in the middle of the line at the same location as \textit{drone1}. The formation performs two rotations, first rotating $120º$ counterclockwise and then $90º$ clockwise.
    
\begin{figure}[htb!]
    \centering
    \includegraphics[width=0.9\linewidth]{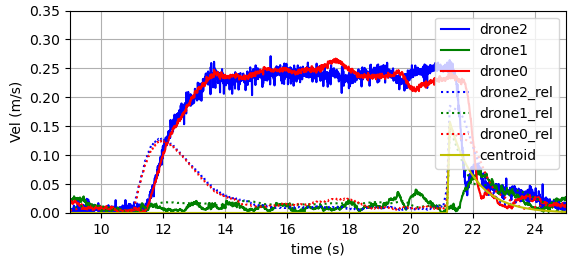}
    \caption{The solid lines are the absolute velocities for the swarm and the three drones. The dotted lines are the relative speeds of the drones in the real z-axis rotation.}
    \label{fig:twist3}
\end{figure}
\begin{figure}[htb!]
    \centering
    \includegraphics[width=0.9\linewidth]{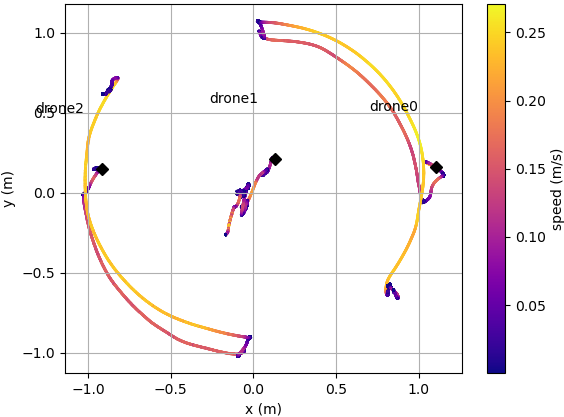}
    \caption{Zenithal view for the real experiment with the z-axis rotation according to the individual velocity of each drone. Color map on the right represents velocities between 0.0 and 0.5 $m/s$.}
    \label{fig:colored3}
\end{figure}

Table \ref{tab:real} summarizes the results obtained. Fig. \ref{fig:colored3} shows the trajectory with the corresponding color map for the speed of the drone. Fig. \ref{fig:twist3} shows each drone speed $||\dot{P}_i^{W}(t)||$ during the experiment, together with their relative speed to the centroid $||\dot{P}_i^{VC}(t)||$ and the centroid speed $||\dot{P}_{VC}^{W}(t)||$. In this case, the $||\dot{P}_{VC}^{W}(t)||$ is closer to zero because there is no translation of the swarm, only a rotation around its z-axis. Since \textit{drone1} is at the same placement, its $||\dot{P}_1^{W}(t)||$ is below $0.05 m/s$, as expected. However, the other drones follow their $P_i^{VC}(t)$ and perform a curvilinear trajectory with a $||\dot{P}_i^{VC}(t)||$ below 0.038 m/s.

\section{CONCLUSIONS AND FUTURE WORK}
This paper presents a new way to structure a swarm by introducing the \textit{FlockingBehavior} algorithm. This algorithm presents a simple way to achieve complex geometric formations that can change during the execution of a flight. It also shows the possibility of creating complex trajectories with multiple agents by assigning only one trajectory to the $VC$. 
The simulated and real experiments carried out show compliance with the Reynolds' rules.
Future work includes the de-centralization of the swarm from the agents. We also plan to develop an algorithm for implementing a complementary strategy, which will deform the formation in response to the environment.
\color{black}
\section*{\small{ACKNOWLEDGMENT}}
\footnotesize{We acknowledge the support of the European European Union through the Horizon Europe Project No. HZ230070443 SHEREC: Safe, Healthy and Environmental Ship Recycling and the Horizon Europe Project No. 101070254 CORESENSE. This work has also been supported by Grant ref. ID2021-127648OBC32 INSERTION funded by MICIU/AEI/10.13039/501100011033 and by FEDER, EU. The work of the second author is supported by the Grant FPU20/07198 of the Spanish Ministry for Universities. The work of the fourth author is supported by the Spanish Ministry of Science and Innovation under its Program for Technical Assistants PTA2021-020671.}

\bibliographystyle{unsrt}
\bibliography{bibliography}
\end{document}